\documentclass[a4paper,fleqn]{cas-dc}

\usepackage[authoryear,longnamesfirst]{natbib}
\usepackage{graphicx}%
\usepackage{multirow}%
\usepackage{amsmath,amssymb,amsfonts}%
\usepackage{amsthm}%
\usepackage{mathrsfs}%
\usepackage[title]{appendix}%
\usepackage{xcolor}%
\usepackage{textcomp}%
\usepackage{manyfoot}%
\usepackage{booktabs}%
\usepackage{algorithm}%
\usepackage{algorithmicx}%
\usepackage{algpseudocode}%
\usepackage{listings}%
\usepackage{graphicx} 
\usepackage{booktabs} 
\usepackage{siunitx} 
\usepackage{float}

\usepackage{tabularx} 
\usepackage{array} 

\def\tsc#1{\csdef{#1}{\textsc{\lowercase{#1}}\xspace}}
\tsc{WGM}
\tsc{QE}
\tsc{EP}
\tsc{PMS}
\tsc{BEC}
\tsc{DE}

\theoremstyle{thmstyleone}%

\theoremstyle{thmstyletwo}%

\theoremstyle{thmstylethree}%

\begin{document}
\let\WriteBookmarks\relax
\def\floatpagepagefraction{1}
\def\textpagefraction{.001}

\shorttitle{Strategic Text Augmentation in NLP Models}

\shortauthors{HT Kesgin et~al.}

\title[mode = title]{Advancing NLP Models with Strategic Text Augmentation: A Comprehensive Study of Augmentation Methods and Curriculum Strategies}

\tnotemark[1,2]

\tnotetext[1]{This study was supported by the Scientific and Technological Research Council of Turkey (TUBITAK) Grant No: 120E100.}


\author[1]{Himmet Toprak Kesgin}\ead{tkesgin@yildiz.edu.tr}
\cormark[1]
\author[1]{Mehmet Fatih Amasyali}\ead{amasyali@yildiz.edu.tr}

\affiliation[1]{organization={Yildiz Technical University, Computer Engineering Depeartment},
    city={Istanbul},
    postcode={34220}, 
    state={Esenler},
    country={Turkey}}

\begin{abstract}
This study conducts a thorough evaluation of text augmentation techniques across a variety of datasets and natural language processing (NLP) tasks to address the lack of reliable, generalized evidence for these methods. It examines the effectiveness of these techniques in augmenting training sets to improve performance in tasks such as topic classification, sentiment analysis, and offensive language detection. The research emphasizes not only the augmentation methods, but also the strategic order in which real and augmented instances are introduced during training. A major contribution is the development and evaluation of Modified Cyclical Curriculum Learning (MCCL) for augmented datasets, which represents a novel approach in the field. Results show that specific augmentation methods, especially when integrated with MCCL, significantly outperform traditional training approaches in NLP model performance. These results underscore the need for careful selection of augmentation techniques and sequencing strategies to optimize the balance between speed and quality improvement in various NLP tasks. The study concludes that the use of augmentation methods, especially in conjunction with MCCL, leads to improved results in various classification tasks, providing a foundation for future advances in text augmentation strategies in NLP.
\end{abstract}


\maketitle

\section{Introduction}\label{sec1}

Text data is everywhere in today's online world and is key to many natural language processing (NLP) applications. It can be found in social media, online reviews, news sites, and academic papers. NLP uses this data for tasks such as sentiment analysis, topic classification, question answering, and chatbots. As industries like finance, healthcare, and marketing begin to use NLP more and more, it's important that we overcome the challenges of dealing with text data. 

One of the most significant challenges is sparsity, which occurs when words or phrases appear only once or twice in the corpus. Another significant challenge is the lack of labeled data. This can limit the performance of supervised learning algorithms, which are essential for data analysis. Deep learning models are becoming more and more successful as the amount of data increases. However, the performance of the models can be limited by the cost of obtaining labeled data. 
This research addresses these challenges by pioneering a method of text augmentation that prioritizes the strategic sequencing and integration of augmented sentences within the training process, with the goal of significantly improving the efficiency and effectiveness of NLP models. 
The study also establishes a comprehensive framework for evaluating text augmentation methods across different NLP tasks, providing new insights into their effectiveness.

The goal of text augmentation is to improve the performance of machine learning models by creating additional training examples and increasing the diversity of the data set. It can prevent overfitting and underperformance on unseen data. Text augmentation techniques include adding, deleting, or replacing words in the text with synonyms \cite{wei2019eda, marivate2020improving}, changing the order of words, adding or deleting punctuation \cite{karimi2021aeda}, back-translation \cite{ma2020data, shleifer2019low}, generative models \cite{qiu2020easyaug, anaby2020not, wang2019classification}, and vector-level augmentation such as random noise insertion \cite{cheung2021modals}, mix-up \cite{zhangmixup}, etc.

Text augmentation has not yet received the same level of attention as image augmentation. This is probably due to the structural differences between text and image data. While image data is continuous, text data is discrete, making it difficult to modify text data without changing its meaning or structure. However, more sophisticated and effective text augmentation methods have been made possible by recent advances in NLP techniques, such as pre-trained language models.

This study conducts a thorough comparative analysis of text augmentation methods across multiple dimensions, including dataset types, augmentation techniques, filtering processes, and augmentation size. This broadened scope of evaluation represents a significant advancement in the field, extending beyond the confines of previous research. Additionally, a novel adaptation of CCL algorithm, termed MCCL algorithm and designed for use with datasets containing artificial examples, is proposed and its effectiveness demonstrated, further contributing to the methodological innovations in NLP.

The strategic sequencing of artificial examples in model training, akin to an advanced form of curriculum learning, is the focus of this research, addressing a critical but underexplored area in NLP.
We investigate how different sequencing strategies of augmented data affect model training, emphasising the importance of order and priority. This targeted exploration requires a comprehensive application of different text augmentation methods, which forms the basis of our experimental investigations.

Our research is driven by the following questions: Can expanding the training set through automatic data augmentation enhance test performance? What are the relative advantages of different automatic data augmentation methods? Is there a method that consistently excels across most datasets? How does augmentation size affect performance? What is the impact of filtering on augmented data? How do different training sequences affect model performance?

These questions not only direct our experimental designs but also aim to provide a comprehensive overview of text augmentation's role in enhancing NLP model performance, offering valuable insights for both researchers and practitioners.

This paper is structured to first review existing augmentation methods, setting the stage with a literature review that underscores the background and challenges of handling textual data for NLP. The methodology section then elaborates on the dataset selection criteria, detailed descriptions of the augmentation methods, and the evaluation framework, including the novel MCCL approach. The experiments section presents a comprehensive evaluation of the augmentation methods, the effect of filtering on the augmented samples, the effect of different training sequences, and an analysis of the execution time of the augmentation methods. The discussion synthesizes these results and highlights their importance for improving NLP model development, and the paper concludes by summarizing the main findings and suggesting avenues for future research.

\section{Literature}\label{sec2}

{
In the field of NLP, a variety of text augmentation techniques have been developed to increase the quality and quantity of training data. 
These methods, which range from simple to sophisticated, serve the common goal of improving the performance of NLP models.

Among these methods, Easy Data Augmentation (EDA) \cite{wei2019eda} presents a simple yet robust data augmentation technique for improving text classification tasks. The EDA method proposes four techniques, namely synonym replacing, random inserting, random replacing, and random deleting. The primary goal of EDA is to improve model performance. The authors provide empirical evidence that demonstrates the effectiveness of EDA in significantly improving the accuracy of the models on the small datasets.

Another similar study, An Easier Data Augmentation (AEDA) \cite{karimi2021aeda}, proposes a novel approach to data augmentation: Adding random punctuation to text. In contrast to EDA, AEDA is easier to implement. It preserves all input information. The authors conduct experiments on five different datasets. They show that using AEDA-augmented data for training yields superior performance compared to EDA-augmented data.

Another type of text augmentation technique that can be used to increase the amount of text is the back-translation technique. Back-translation is the process of translating a text from its original language into a different language, and then translating the resulting translation back into the original language. It has been demonstrated that the use of back-translation can lead to a reduction in overfitting and an improvement in model performance \cite{ma2020data, shleifer2019low}.

In recent years, word embedding techniques such as word2vec \cite{mikolov2013efficient}, FastText \cite{bojanowski2017enriching}, GloVe \cite{pennington2014glove}, and Mask Filling language models \cite{devlin2018bert} have gained popularity as effective methods for text augmentation. These techniques aim to enhance the quality of training data by replacing a word with one of its closest embeddings. Compared to the traditional synonym replacement technique, word embedding-based augmentation provides a more diverse set of augmented data. Moreover, for models like BERT, the use of Mask Filling language models allows for the random replacement of words with a Mask, which is then filled with model predictions. Overall, the use of word embedding techniques in text augmentation is an effective strategy to enhance the quality and quantity of training data.

Recent advances in NLP have led to the development of powerful models, such as Generative Pretrained Transformers (GPTs) \cite{radford2019language}, that can generate coherent, semantically meaningful text. The ability of these transformer-based models to generate entirely new sentences and paragraphs from an existing dataset has opened up new possibilities for text augmentation. Researchers can generate a much larger and more diverse set of synthetic data to train their models by feeding the original text data into the transformer models. In this context, transformer-based text augmentation methods have shown promising results and are becoming increasingly popular for improving the performance of various NLP tasks.

Generative Adversarial Networks (GANs)  \cite{goodfellow2020generative} originally developed for images have become a significant innovation in text augmentation. GANs now generate synthetic text for NLP, supporting tasks such as sentiment analysis, language generation in low-resource environments, and detection of complex patterns such as hate speech and fraud \cite{silva2023text}. GANs help overcome data scarcity by enriching training datasets with diverse, realistic text. However, there are challenges to using GANs in NLP, particularly in generating authentic, language-compliant text.

The Iterative Mask Filling (IMF) augmentation \cite{10.1007/978-3-031-50920-9_35}, utilizing the Fill-Mask feature of BERT, marks a significant advancement. This method iteratively masks and replaces words in sentences, maintaining the original text's meaning while significantly diversifying the training data. Proven effective in various NLP tasks, especially in topic classification, IMF outperforms traditional text augmentation techniques, highlighting the potential of sophisticated language models in enhancing data quality for NLP.

Another notable advancement in text augmentation is the use of autoencoder-based augmentation, particularly through Conditional Variational Autoencoders combined with Transformer models, which marks a significant step forward in NLP \cite{bilici2022transformers}. This method adeptly generates class-specific, syntactically and semantically rich synthetic sentences, thereby enhancing the diversity of the dataset. It outperforms traditional techniques in various NLP tasks, demonstrating its potential to efficiently improve model training and effectiveness.

Data augmentation techniques are not limited to the modification of existing samples. Instead, the feature space can also be augmented by adding random noise \cite{cheung2021modals}, combining two samples using linear interpolation \cite{zhangmixup}, or activating dropout \cite{srivastava2014dropout} during the construction of the model vector space to generate different representations for the same instances. Activating dropout during the construction of the model vector space can further enhance the diversity of the feature space and improve the performance of the model. Therefore, leveraging these techniques can increase the amount and diversity of the training data, leading to better performance of the models.

\begin{table*}[!htb]
\centering
\caption{Text Augmentation Techniques}
\label{tab:TextAugmentationTechniques}
\begin{tabularx}{\textwidth}{|>{\raggedright\arraybackslash}p{2.5cm}|>{\raggedright\arraybackslash}X|>{\centering\arraybackslash}p{2cm}|>{\centering\arraybackslash}p{2cm}|}
\hline
\textbf{Method} & \textbf{Description} & \textbf{Resource Intensity} & \textbf{Type} \\ 
\hline
EDA & Applies operations like synonym replacement, random insertion, swapping, and deletion. & Low & Text-based \\ 
\hline
AEDA & Extends EDA by adding random punctuation to text, preserving all input info. & Low & Text-based \\ 
\hline
Back Translation & Translates text to a different language and back, reducing overfitting. Requires translation model. & High & Text-based \\ 
\hline
Word2Vec \& FastText & Replaces words with synonyms based on embeddings, enhancing quality. Requires pre-trained embeddings. & Medium & Text-based \\ 
\hline
IMF & Iteratively masks and predicts words in sentences, using context for diversity. Utilizes BERT's Fill-Mask feature. & Medium & Text-based \\ 
\hline
GPT & Generates new text based on existing samples, providing diverse expansions. Utilizes GPT models. & High & Text-based \\ 
\hline
Autoencoder-Based & Generates class-specific, syntactically, and semantically rich synthetic sentences. Uses Conditional Variational Autoencoders and Transformers. & High & Text-based \\ 
\hline
Random Noise & Adds random noise to embedding vectors, increasing robustness. & Low & Vector-based \\ 
\hline
Mixup & Combines examples and labels linearly, encouraging generalization. & Low & Vector-based \\ 
\hline
Dropout Activation & Applies dropout during augmentation for ensemble-like diversity. & Low & Vector-based \\ 
\hline
\end{tabularx}
\end{table*}

Table \ref{tab:TextAugmentationTechniques} provides a summary of these text augmentation techniques, detailing their descriptions, resource intensity, and whether they are vector or text-based, illustrating the breadth and diversity of methods available for enhancing NLP model performance.

As a result, text augmentation techniques have been widely investigated in the field of NLP to improve the quality and quantity of training data. In general, the use of text augmentation techniques offers a promising approach to improve the performance of NLP models and to overcome the challenges of limited training data.
}

\section{Method}\label{sec3}

We evaluated various text augmentation methods on ten diverse datasets for multiple NLP tasks, including topic classification, sentiment analysis, and offensive language detection. Each dataset originated from a different domain, with unique tasks and class numbers, ensuring a wide-ranging evaluation. table \ref{tab:datasets_summary} provides an overview of the selected datasets, highlighting their varying class numbers and tasks for comprehensive coverage.

\begin{table*}[ht]
\centering
\caption{Summary of datasets used in the study}
\label{tab:datasets_summary}
\begin{tabular}{|l|l|l|}
\toprule
\textbf{Dataset} & \textbf{Classes} & \textbf{Task} \\
\midrule
ag news \cite{Zhang2015CharacterlevelCN} & 4 & Topic Classification \\
news \cite{misra2022news} & 10 & Topic Classification \\
twitsent \cite{go2009twitter} & 2 & Sentiment Analysis \\
airline \cite{airline} & 3 & Sentiment Analysis \\
imdb  \cite{imdb} &  2 & Sentiment Analysis \\
yahoo \cite{yahoo} & 10 & Topic Classification \\
rotten\_tomatoes \cite{rotten} & 2 & Sentiment Analysis \\
tweet\_eval\_offensive \cite{zampieri2019semeval} & 2 & Offensive Language Detection \\
dbpedia\_14 \cite{zhang2015character} & 14 & Topic Classification \\
emotion \cite{saravia-etal-2018-carer} & 6 & Emotion Recognition \\
\bottomrule
\end{tabular}
\end{table*}

\subsection{Dataset Selection Criteria}

The selection of datasets was a critical step in our methodology. We aimed to cover a broad spectrum of NLP tasks and domains, ensuring that our findings would be widely applicable and informative. Here, we detail the criteria behind our dataset selection and the rationale for the preprocessing steps we undertook.

In our methodology, we introduced a preprocessing step that converted all sentences in the datasets to lowercase and limited them to the first 300 characters. These datasets were then partitioned into a training set and a test set using stratified sampling with 1000 and 4000 samples, respectively. A key divergence in this process was applied to the twitsent dataset, where we selected only 500 samples for the training set. Our initial investigations showed that selecting 1000 samples this particular dataset, followed by the addition of another 1000 real samples, did not result in a statistically significant difference. In contrast, when we limited the twitsent training set to 500 samples, we observed that the inclusion of real samples then elicited statistically significant variations. For the remaining datasets, we remained consistent with our original selection of 1000 samples. Adding another 1000 samples to these datasets indeed yielded significant results. These findings emphasized the potential impact of well-curated augmented samples.

Our decision to work with smaller training datasets was deliberate. This approach was informed by our understanding that an unwarranted expansion of already large datasets could potentially result in negligible performance gains. By using a more focused subset of the data, we were better equipped to evaluate the performance of various text augmentation methods, thus drawing more insightful conclusions about their effectiveness across a range of NLP tasks.

\subsection{Text Representation and Modeling Techniques}

In this section, we delve into the text representation and modeling techniques employed in our study. We explain our choice of techniques, including why we specifically opted for the BERT model despite the existence of various alternatives, highlighting their relevance to our study's objectives.

Various text representation and modeling techniques, such as Term Frequency-Inverse Document Frequency (TF-IDF) and Word2Vec for generating text vectors, as well as Recurrent Neural Networks (RNN) and Long Short-Term Memory (LSTM) networks for processing sequential text data, can often improve their performance in tasks such as text classification when combined with text augmentation techniques.

Similarly, smaller transformer models, which have fewer parameters than their larger counterparts, can also show improved performance with these techniques, despite lower baseline results. 
In this study, however, we specifically chose to use the BERT model \cite{devlin2018bert}. 
This decision was made because BERT has consistently demonstrated superior performance in vanilla applications, thus serving as an effective benchmark. 
While the advances with alternative representation methods and smaller models are valuable, our goal was to optimize performance where it was already strongest. 
Therefore, our focus was on exploring the potential of various text augmentation methods to improve the already high performance of the BERT model. 
In our process, we transformed each text sample into a 768-dimensional vector using the BERT representation.

We used a number of state-of-the-art text augmentation methods to create augmented training samples. 
These included back-translation, IMF with BERT and TinyBERT, dropout, mix-up, noise insertion, GPT-2, random deletion, random insertion, random swapping, synonym replacement, Word2Vec replacement, and AEDA. 
Our primary goal was to provide a comprehensive comparison of a large and diverse set of augmentation methods. 

\subsection{Text Augmentation Methods}

Our study employed a diverse array of text augmentation methods, each chosen for its potential to enhance the model's performance on various NLP tasks. This section provides an overview of these methods, their implementation, and the rationale behind their selection.

The back translation method used a Helsinki NLP model \cite{TiedemannThottingal:EAMT2020} to translate text to and from German and English. 
IMF involved masking a random set of tokens in a sentence and then predicting the masked tokens using the BERT or TinyBERT model \cite{bhargava2021generalization}. 
The dropout method enabled dropout to generate multiple representations of the same text, while mixup generated new text by interpolating pairs of randomly selected samples from the training set. 
Noise insertion added random noise to the generated BERT vectors.
In another method for text augmentation, we used a pre-trained GPT-2 language model.
In this method, we split the text into two parts, took the first part, generated the second part using the GPT-2 model, and combined the two parts to create an augmented sentence.
The random deletion method involved randomly deleting words from the text, while the random insertion method involved randomly inserting words into the text.
In the random swapping method, we randomly swapped adjacent words in the text.
In the random deletion method, words were randomly deleted from the text, while in the random insertion method, words were randomly inserted into the text.
For the random swapping method, we randomly swapped adjacent words in the text.
For the synonym replacement method, we used WordNet \cite{miller1995wordnet} to replace words in the text with their synonyms. Meanwhile, in the Word2Vec replacement method, we replaced words in the text with others that have similar vector representations according to a pre-trained Word2Vec model.
Finally, the AEDA method added random punctuation to the text.

Guided by the EDA study, we set the alpha parameter to 0.1 for Random Deletion, Random Insertion, Random Swapping, Synonym Replacement, and also for Word2Vec Replacement, as these were found to produce the best results in that study. For the IMF and Mixup methods, we adopted the hyperparameters suggested in their respective papers. Back translation was implemented using the default hyperparameters of the translation model. For the noise addition process, we added random noise from a distribution with a mean of 0 and a standard deviation of 1. In the GPT-2 method, before text generation, we randomly split the text in half between 80 and 120 characters. Finally, both the EDA and AEDA methods were applied using the default hyperparameters specified in their respective articles.

Each method augmented the initial training samples by 300\% allowed for a comprehensive evaluation of their effectiveness.

\subsection{Experimental Setup and Evaluation Metrics}

To assess the effectiveness of our text augmentation methods, we developed a rigorous experimental setup, detailed in this section. We outline our evaluation metrics and training procedures, explaining how these choices align with the overall goals of our study.

To evaluate the effectiveness of the text augmentation methods, we converted each sample into a BERT vector and used a neural network with two hidden layers, each containing 64 neurons. The ReLU activation function was employed in these layers, and the softmax activation function was used in the last layer, corresponding to the number of classes in the dataset. The neural network was trained using the Adam optimization algorithm with default parameters, and performance was evaluated using the accuracy metric. Since the samples in the datasets were approximately evenly distributed, accuracy was an appropriate success criterion for our study.

To establish a baseline for comparison, we trained the neural network using the vanilla method, without any augmentation. We then applied each text augmentation method to the training data and retrained the neural network using the augmented data.

To ensure robustness, we ran each method 20 times, calculated the mean accuracy and standard deviation, and used a two-tailed t-test to determine whether the difference between the mean accuracy of each method and the vanilla method was statistically significant at a significance level of p = 0.05.

We individually evaluated each method using both 100\% and 300\% of the original sample size and analyzed their impact on performance metrics. Additionally, we filtered samples with low loss values and explored the effects of presenting augmented and original data to the model in different sequences. This comprehensive approach allowed us to emphasize the key aspects of our evaluation. Our experimental approach provides a comprehensive evaluation of text augmentation methods for NLP tasks and aims to advance the field by providing insights to improve model performance. 

In conclusion, our methodological choices were carefully designed to provide a comprehensive evaluation of text augmentation methods in NLP. By focusing on a range of datasets and augmentation techniques, we aimed to offer insights that could significantly contribute to advancements in the field. Future work may explore the application of these methods in other NLP tasks and incorporate emerging techniques for potentially enhanced performance.

\section{Experiments}
In this section, we present the results of our extensive experimental evaluation of various text augmentation methods. 
Our goal was to evaluate their potential for improving the performance of the base (vanilla) model. 

\subsection{Evaluation of Text Augmentation Methods}

For a thorough analysis, we organized the methods according to augmentation rates of 100\% and 300\%.
The experimental design involved running each model configuration 20 times to ensure the reliability of the results. 
This approach was taken to address our research question regarding the impact of different augmentation rates on model performance.
To test the effectiveness of the different augmentation methods, we used the performance of the non-augmented (vanilla) model as a benchmark. 
We considered improvements to be statistically significant if the t-test yielded a value of \(p = 0.05\). The results are summarized in Table \ref{tab:results1}, which reports the performance metrics for all methods.

\begin{table}[h]
\centering
\caption{Results Displaying Method, Augmented Data Size, and Number of Wins and Losses Across 10 Datasets}
\label{my-label}
\begin{tabular}{|lccc|}
\toprule
\textbf{Method} & \textbf{Aug Rate} & \textbf{Wins} & \textbf{Losses} \\ 
\midrule
aeda     & 100\%  & 0  & 0    \\ 
aeda     & 300\%  & 1  & 2    \\ 
back\_tr  & 100\%  & 0  & 2    \\ 
back\_tr  & 300\%  & 0  & 8    \\ 
Bert-IMF & 100\%  & 0  & 8    \\ 
Bert-IMF & 300\%  & 0  & 9    \\ 
dropout  & 100\%  & 0  & 4    \\ 
dropout  & 300\%  & 0  & 5    \\ 
gpt2     & 100\%  & 1  & 0    \\ 
gpt2     & 300\%  & 1  & 3    \\ 
mixup    & 100\%  & 1  & 2    \\ 
mixup    & 300\%  & 1  & 8    \\ 
noise    & 100\%  & 3  & 1    \\ 
noise    & 300\%  & 0  & 2    \\ 
rd       & 100\%  & 1  & 2    \\ 
rd       & 300\%  & 1  & 6    \\ 
ri       & 100\%  & 3  & 2    \\ 
ri       & 300\%  & 0  & 5    \\ 
rs       & 100\%  & 3  & 3    \\ 
rs       & 300\%  & 0  & 6    \\ 
sr       & 100\%  & 2  & 4    \\ 
sr       & 300\%  & 1  & 6    \\ 
Tiny-IMF & 100\%  & 1  & 8    \\ 
Tiny-IMF & 300\%  & 1  & 9    \\ 
w2v      & 100\%  & 0  & 0    \\ 
w2v      & 300\%  & 1  & 5    \\ 
\bottomrule
\end{tabular}
\label{tab:results1}
\end{table}

Our results provide important insights into the effectiveness of different text augmentation methods compared to the vanilla model. 
This directly addresses our research question on the relative advantages of different augmentation methods.
Importantly, while certain methods provided some improvement in model performance, none consistently demonstrated clear superiority over the vanilla model.

In particular, GPT-2 and Noise proved to be slightly more effective than the other augmentation methods, while Back Translation and IMF showed limited effectiveness. 
This finding contributes to our understanding of how specific methods, such as GPT-2 and Noise, can be more conducive to certain tasks, addressing another aspect of our research question on the variability of these methods across different datasets. 
The slightly improved performance of GPT-2 could be attributed to its extensive pre-training dataset, which provides additional information beyond the provided dataset and its sentences. Similarly, the Noise method likely benefits from acting as a regularization technique, reducing overfitting and thus improving performance.

However, it is crucial to contextualize that the marginal benefit observed with GPT-2 is not necessarily universal, as it depends on the match between pre-training and downstream tasks. Moreover, the small performance advantage conferred by GPT-2 and noise does not clearly translate into superior performance across all tasks and data sets.

One observation from our study relates to the effect of the augmentation rate on the performance of these methods. In general, increasing the augmentation rate from 100\% to 300\% resulted in more losses compared to the vanilla model. 
This observation is critical in understanding the optimal balance in augmentation quantity, which is essential to one of our research questions.
This finding suggests that augmented sentences could potentially introduce noise, meaning that a higher augmentation rate may not always result in better performance. Thus, the notion that more augmented data would always lead to better performance is not necessarily supported by our results.

\subsection{Impact of Filtering on Augmented Data}

\begin{table}[ht]
\centering
\caption{Best results for each augmentation method}
\begin{tabular}{l S[table-format=4.0] S[table-format=1.2] S[table-format=1.0] S[table-format=1.0]}
\toprule
\textbf{Augmentation} & \textbf{Size} & \textbf{Filter} & \textbf{Win} & \textbf{Loss} \\
\midrule
aeda     & 100\% & 0.5  & 0 & 3 \\
back\_tr  & 100\% & 0.75 & 0 & 2 \\
bert     & 100\% & 0.5  & 0 & 6 \\
dropout  & 100\% & 0.5  & 1 & 3 \\
gpt2     & 100\% & 0.75 & 0 & 3 \\
mixup    & 100\% & 0.5  & 1 & 2 \\
noise    & 100\% & 0.5  & 1 & 0 \\
rd       & 100\% & 0.5  & 0 & 2 \\
ri       & 100\% & 0.75 & 1 & 3 \\
rs       & 100\% & 0.75 & 2 & 3 \\
sr       & 100\% & 0.5  & 1 & 4 \\
tiny     & 100\% & 0.5  & 1 & 7 \\
w2v      & 100\% & 0.75 & 0 & 3 \\
\bottomrule
\end{tabular}
\label{tab:filter}
\end{table}

In our investigation, we also examined the effect of filtering on the augmented samples. Given the possibility of noise in the augmented data, we aimed to determine whether strategic filtering based on loss values could enhance the performance of augmented models in comparison to the vanilla model. This examination directly relates to our research question about optimizing the quality of augmented data for improved performance in NLP models.

To conduct this, we initially trained the vanilla model without any augmentation and computed the loss values for each augmented sample. Subsequently, we applied two filtering criteria, retaining either the lower 50\% or 75\% of the loss values. From these, we added the most successful result from the 50\% and 75\% filtered data. These results are shown in Table \ref{tab:filter}.

Our findings highlight that even with such selective filtering, the overall performance of the augmentation methods does not consistently outperform the vanilla model.  
In fact, while some methods such as Dropout and Mixup showed improved results in specific instances, the general trend did not indicate a notable performance improvement due to filtering. 
This outcome provides an important insight into our research question regarding the effectiveness of data quality control in text augmentation. It suggests that merely filtering based on loss values may not be sufficient to ensure enhanced performance of NLP models.
This reinforces the notion that the introduction of noise through augmented data cannot always be mitigated by filtering strategies, and the non-augmented vanilla model often remains a robust baseline for comparison.

\subsection{Effect of Different Training Sequences on Model Performance}

However, it is important to consider that in these experiments, augmented samples were treated the same way as real samples during model training. The effectiveness of these sequencing strategies in enhancing model performance was evaluated to directly respond to our research question on the impact of training sequence order in the context of augmented data. We suspect that acknowledging the difference between real and augmented samples and presenting them to the model in different orders might have a significant impact on the model's performance. In our further experiments, we explore various strategies for feeding augmented samples to the model, such as presenting artificial data first followed by real data, or vice versa. By examining the effect of presenting augmented samples in different orders, we aim to determine whether treating augmented samples differently can lead to better results and help improve the performance of NLP models.

Hence, In addition to the previously mentioned augmentation methods, we also employed various sequencing strategies for presenting augmented samples during model training. These strategies include: Curriculum Learning (CL) \cite{bengio2009curriculum}, Anti-Curriculum Learning (AntiCL), Cyclical Curriculum Learning (CCL) \cite{kesgin2023cyclical}, Random Curriculum (RandCL), Modified CCL (MCCL), Augmented Data First (ADF), augmented data after (ADA), augmented data in the middle (ADM).

To briefly explain these methods, CL involves sorting data from easy to difficult, presenting the model with easier data first and more challenging data later. AntiCL reverses this process, exposing the model to difficult data first, followed by easier examples. In RandCL, the data is ordered randomly but provided to the model incrementally, distinguishing it from the vanilla method where all data is presented simultaneously.

CCL applies a cyclical approach, alternating between curriculum-based and vanilla training, with examples ordered from easy to difficult. In the original CCL algorithm, the vanilla model is trained using the entire dataset in these case both real and augmented data, to obtain easiness scores for instances in the dataset, which then guide the CCL process. In contrast, MCCL only uses real data during the initial vanilla model training, instead of using both real and augmented data, as in the original CCL algorithm. However, it still calculates easiness scores for both real and augmented data samples. Lastly, MCCL performs the training by combining both real and augmented examples and using these easiness scores. This distinction highlights the difference between the original CCL algorithm and the MCCL method, with the latter focusing solely on real data during the initial vanilla model training.


\begin{algorithm}
\caption{MCCL Training Algorithm}
\begin{algorithmic}[1]
\Require
\State $D_{real}$: Real training dataset (X: Input, y: Output)
\State $D_{aug}$: Augmented training dataset (X: Augmented Input, y: Output)
\State $M_1$: Model for score generation
\State $M_2$: Model to be trained
\State $T$: Number of training epochs
\State $ip$: Initial dataset size percentage
\State $fp$: Final dataset size percentage
\State $\alpha$: Speed of cycle adjustment
\Statex

\Procedure{MCCL\_Train}{}
    \State Train $M_1$ with $D_{real}$ to generate scores
    \State Combine $D_{real}$ and $D_{aug}$ into $D_{combined}$
    \State Calculate easiness scores for $D_{combined}$ using $M_1$
    \State Initialize $M_2$
    \For{each epoch $t$ from 1 to $T$}
        \State Determine subset $D_{subset}$ from $D_{combined}$ based on scores and cycle size $S[t]$
        \State Update $M_2$ with $D_{subset}$ for the current epoch
    \EndFor
    \State \textbf{return} Trained model $M_2$
\EndProcedure
\end{algorithmic}
\end{algorithm}

MCCL algorithm is given in Algorithm 1. In MCCL, the initial training phase focuses solely on real data to preserve the integrity of the model's understanding of genuine patterns. Subsequently, a blend of real and augmented data, selected based on easiness scores, is used to enrich the training process and enhance model robustness. This method leverages the strengths of curriculum learning by gradually introducing complexity and diversity into the training regimen.

In ADF sequencing strategy, the model is first exposed to the augmented data. Once the model has been trained on this data, the original, non-augmented dataset is introduced for further training. ADA strategy reverses the order of the previous approach. The model is initially trained on the original, non-augmented dataset, and then the augmented or artificially generated data is provided for additional training. In ADM method, the training process starts with a portion of the original, non-augmented dataset. Then, the augmented or artificially generated data is introduced, followed by the remaining portion of the original dataset to complete the training process.

This comprehensive analysis aims to provide clear answers to our research questions about the adaptability and efficiency of various sequencing strategies when combined with text augmentation techniques.

\begin{table}[h]
\centering
\renewcommand{\arraystretch}{1.3}
\caption{Summary of sequencing strategies and their corresponding implementations}
\begin{tabular}{|l|l|}
\hline
\textbf{Method} & \textbf{Implementation} \\ \hline
Vanilla & \texttt{Model.fit(x\_train, y\_train)} \\ \hline
ADF & \texttt{Model.fit(x\_aug, y\_aug);} \\
 & \texttt{Model.fit(x\_train, y\_train)} \\ \hline
ADM & \texttt{Model.fit(x\_train, y\_train);} \\
 & \texttt{Model.fit(x\_aug, y\_aug);} \\
 & \texttt{Model.fit(x\_train, y\_train)} \\ \hline
ADA & \texttt{Model.fit(x\_train, y\_train);} \\
 & \texttt{Model.fit(x\_aug, y\_aug)} \\ \hline
CL & \texttt{Model.fit(x\_easy, y\_easy);} \\
 & \texttt{Model.fit(x\_mixed, y\_mixed)} \\ \hline
RandCL & \texttt{Model.fit(x\_sub, y\_sub);} \\
 & \texttt{Model.fit(x\_mixed, y\_mixed)} \\ \hline
AntiCL & \texttt{Model.fit(x\_hard, y\_hard);} \\
 & \texttt{Model.fit(x\_mixed, y\_mixed)} \\ \hline
CCL & \texttt{Model.fit(x\_easy, y\_easy);} \\
 & \texttt{Model.fit(x\_mixed, y\_mixed);} \\
 & \texttt{Model.fit(x\_easy, y\_easy)} \\ \hline
MCCL & \texttt{Model.fit(x\_easy, y\_easy);} \\
 & \texttt{Model.fit(x\_mixed, y\_mixed);} \\
 & \texttt{Model.fit(x\_easy, y\_easy)} \\ \hline
\end{tabular}
\label{tab:methods}
\end{table}

Table \ref{tab:methods} provides a concise summary of the sequencing strategies discussed. In this table, the terms x\_train and y\_train correspond to the real instances and their respective labels. Similarly, x\_aug and y\_aug denote the augmented instances and their respective labels. Additionally, x\_mixed and y\_mixed are used to represent a combination of the real and augmented instances and their labels. In the same context, x\_easy and y\_easy denote low-loss samples and their labels derived from vanilla training, while x\_sub and y\_sub denote a random subset of samples and their respective labels. Finally, x\_hard and y\_hard refer to high-loss samples and their vanilla-trained labels. Each of these terms helps to describe different aspects and results of the applied sequencing strategies.

It's important to note that although the sequencing strategies differ in the order and presentation of the training data, each strategy uses the same total number of training epochs. This ensures a fair and balanced comparison between the methods, and allows us to attribute any observed differences in performance to the sequencing strategies themselves, rather than to variations in training duration. Thus, the table \ref{tab:methods} provides a convenient overview of the implementation of each method, and provides an easy reference for understanding and comparing these techniques.

\begin{table}[!ht]
\caption{Top 3 results for each method}
\centering
\renewcommand{\arraystretch}{1.3}
\begin{tabular}{|l|l|c|c|}
\toprule
\textbf{Algorithm} & \textbf{Augmentation} & \textbf{Win} & \textbf{Loss} \\
\midrule
Vanilla             & aeda    & 0 & 0 \\
Vanilla             & noise   & 3 & 1 \\
Vanilla             & w2v     & 0 & 0 \\ \hline
ADF          & aeda    & 1 & 2 \\
ADF           & mixup   & 1 & 0 \\
ADF           & w2v     & 1 & 1 \\ \hline
ADM          & aeda    & 3 & 0 \\
ADM          & gpt2    & 1 & 0 \\
ADM          & w2v     & 2 & 0 \\ \hline
ADA          & back\_tr & 1 & 0 \\
ADA         & dropout & 2 & 0 \\
ADA         & w2v     & 1 & 0 \\ \hline
CL          & dropout & 1 & 3 \\
CL          & gpt2    & 0 & 2 \\
CL          & sr      & 2 & 3 \\ \hline
RandCL   & rd      & 1 & 8 \\
RandCL   & ri      & 1 & 8 \\
RandCL   & sr      & 1 & 8 \\ \hline
AntiCL    & rd      & 0 & 8 \\
AntiCL    & ri      & 2 & 8 \\
AntiCL    & rs      & 2 & 8 \\ \hline
CCL            & aeda    & 8 & 0 \\
CCL            & dropout & 7 & 1 \\
CCL            & gpt2    & 6 & 0 \\ \hline
MCCL     & aeda    & 9 & 0 \\
MCCL     & gpt2    & 10 & 0 \\
MCCL     & ri      & 9 & 0 \\
\bottomrule
\end{tabular}
\label{tab:top_results_method}
\end{table}

\begin{figure*}[!ht]
\centering
\includegraphics[width=\textwidth, height=0.5\textheight]{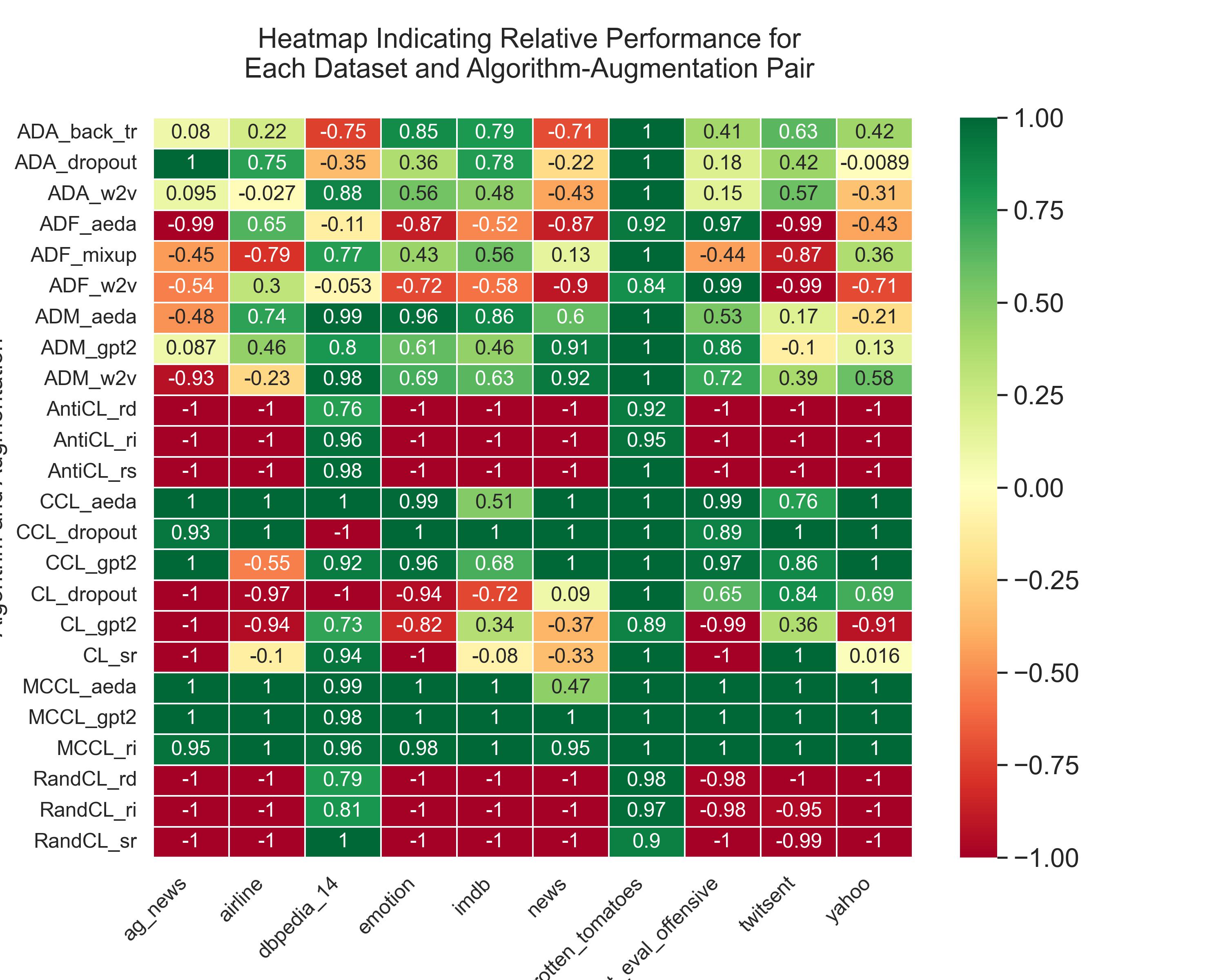}
\caption{Heatmap Indicating Relative Performance for Each Dataset and Algorithm-Augmentation Pair}
\label{heatmap}
\end{figure*}

Table \ref{tab:top_results_method} and Figure \ref{heatmap} presents the 3 most successful examples for each training algorithm. We have focused on these results since there are numerous possible combinations of augmentation methods and training algorithms. Our experiments demonstrate the potential benefits of using various augmentation methods and sequencing strategies to improve the performance of NLP models. The results from Table \ref{tab:top_results_method} and Figure \ref{heatmap} offer insights into our research question about the effectiveness of different training sequences in combination with text augmentation. The top-performing combinations for each training algorithm yielded impressive results, with some cases showing a gain of up to 10 compared to the vanilla method.

The heatmap representation depicted in Figure \ref{heatmap} was designed to provide a graphical interpretation of the results. For ease of interpretation, we subtracted all p-values from one. Subsequently, to highlight the methods that underperformed compared to the vanilla baseline, we assigned them a negative value by multiplying by -1. This way, examples that win against vanilla are close to one, those that perform similarly to vanilla are close to zero, and those that perform worse than vanilla are close to -1. This approach allowed us to visually display the results. We plotted the heatmap of all datasets with the training algorithm augmentation pair that had the top 3 results.

These findings indicate that the training sequence, when used strategically in conjunction with text augmentation, can have a substantial impact on model performance, thereby providing an affirmative response to our research question about training methodologies.

Figure \ref{heatmap} also provides insight into the efficacy of different augmentation methods across various datasets. For instance, while the majority of methods demonstrate significant effectiveness on the dbpedia\_14 and rotten\_tomatoes datasets, it proved challenging to enhance performance on the twitsent and ag\_news datasets. This underscores the importance of tailoring augmentation strategies to the specific characteristics of the dataset at hand.

We observed that vanilla training with non-ranked augmented samples did not yield better results than the standard vanilla method, which does not use augmentation. However, when the MCCL and CCL algorithms were applied in conjunction with various augmentation methods, we noticed substantial performance improvements. Figure \ref{heatmap} highlights these results.

In order to further investigate the contribution of the CCL algorithm to the performance improvements observed, we conducted a comparison between the CCL algorithm using only real examples and the vanilla training using only real examples. The results indicated that the CCL algorithm outperformed the vanilla method with a score of 7 wins and 0 losses. When we compared the performance of the CCL algorithm with augmented examples against the vanilla training with only real examples, the CCL with the AEDA method achieved 8 wins and 0 losses. Only in this case did CCL with augmented samples outperform CCL with real samples. However, when augmentation techniques were applied in conjunction with the MCCL algorithm, the performance improved even further, resulting in a score of 10 wins and 0 losses.

To examine the effect of augmentation independent of the CCL improvement, we compared CCL and MCCL trained with augmented examples with CCL trained with only real examples without augmentation. While CCL trained with augmented examples didn't show much superiority over CCL trained with non-augmented examples, MCCL trained with augmented examples yielded results of 6 wins 1 loss with AEDA, 4 wins 1 loss with GPT2, and 6 wins 2 losses with rs. This demonstrates that MCCL with augmentation methods can even outperform CCL without augmentation, indicating that the success is not solely due to the training algorithm but also due to the combined use of both.

Our experiments may be subject to certain limitations and biases, stemming from our choice of using 1,000 samples from each dataset, employing BERT vectors as sentence representations, and training these vectors with a neural network. The generalizability of our findings might be influenced by these choices, and future research could explore the impact of different augmentation methods on a variety of NLP tasks and model architectures by considering alternative representation methods or architectures, which may yield different results.

In summary, our experiments reveal that the choice of training algorithm and augmentation techniques can play a critical role in achieving optimal performance in NLP tasks. By systematically exploring different training sequences, we have provided insights into how these methods can be optimized to enhance the performance of NLP models, addressing one of our core research questions. CCL and MCCL algorithms, when combined with appropriate augmentation methods, have demonstrated significant improvements compared to the vanilla training method. By carefully selecting and combining these methods, researchers and practitioners can potentially enhance the performance of their NLP models on various tasks and datasets. This investigation into different training sequences with augmented data contributes significantly to our understanding of how to optimize NLP model training, directly answering one of our pivotal research questions.

\subsection{Comparative Analysis of Augmentation Methods Execution Time}

\begin{table}[h]
\caption{Comparison of execution times for different text augmentation strategies}
\centering
\renewcommand{\arraystretch}{1.3}
\begin{tabular}{|c|c|}
\hline
\textbf{Method} & \textbf{Execution Time (ms)} \\
\hline
back\_tr & 2,385,000 \\
\hline
sr & 2.26 \\
\hline
ei & 0.03 \\
\hline
rs & 0.03 \\
\hline
rd & 0.03 \\
\hline
aeda & 0.03 \\
\hline
w2v & 145,000 \\
\hline
gpt2 & 2,477,000 \\
\hline
tiny-imf & 216,000 \\
\hline
bert-imf & 2,173,000 \\
\hline
dropout & 152,000 \\
\hline
noise & 0.008 \\
\hline
mixup & 0.007 \\
\hline
\end{tabular}
\label{tab:time}
\end{table}

In this section, we compare various text augmentation methods in terms of their execution time on an average PC and GPU. The results indicate that some methods, such as sr, ri, rs, rd, and aeda are relatively fast, taking only a few milliseconds to complete. On the other hand, methods like bact\_tr, w2v, gpt2, Tiny-IMF, Bert-IMF, and dropout require significantly more time, ranging from a few hundred thousand to a couple million milliseconds. The noise and mixup methods are the fastest, completing in under a millisecond. The table \ref{tab:time} summarizes these findings.

From these results, we can infer that the choice of text augmentation method may greatly impact the overall time required for data preprocessing. Therefore, it is crucial to consider the trade-offs between the speed and the potential quality improvement offered by each method when selecting an appropriate text augmentation technique for a given task.

In this study, we focus on offline text augmentation. It is important to note that methods with longer execution times are primarily suitable for offline augmentation due to their significant time requirements. Using these methods for online augmentation may not be feasible or efficient. In contrast, faster methods can potentially be used for both offline and online data augmentation, depending on the specific use case and performance requirements.

\section{Conclusion}

Our extensive research into text augmentation in NLP models yielded several significant insights. Below are the key findings categorized under specific subheadings for clarity:

\subsection{Effectiveness of Specific Augmentation Methods}
\begin{itemize}
\item GPT-2 and Noise marginally outperformed methods like Back Translation and IMF, especially when used without supplementary ranking or training algorithms.
\item No single augmentation method consistently surpassed others across all datasets, suggesting the importance of choosing augmentation techniques based on specific context and dataset requirements.
\end{itemize}

\subsection{Impact of Augmentation Rate}
\begin{itemize}
\item Increasing augmentation rate did not always correlate with improved performance. In certain scenarios, a higher augmentation rate resulted in decreased performance compared to the vanilla model.
\item This finding challenges the prevailing notion that more augmented data necessarily leads to better model performance.
\end{itemize}

\subsection{Role of Filtering in Augmented Data}
\begin{itemize}
\item Filtering augmented data based on loss values did not consistently improve performance over the vanilla model. While Dropout and Mixup showed improvements in certain cases, the overall benefit of filtering remained unclear.
\end{itemize}

\subsection{Influence of Training Sequences}
\begin{itemize}
\item Different sequencing strategies, such as ADF, ADA, and MCCL, significantly influenced model performance. MCCL in particular showed substantial performance enhancement over vanilla training.
\end{itemize}

\subsection{Execution Time Considerations}
\begin{itemize}
\item Execution times for text augmentation methods varied widely, affecting their suitability for different applications. Faster methods are more suitable for online augmentation, while slower ones are better suited for offline use.
\end{itemize}

\subsection{Future Research Directions}
\begin{itemize}
\item Future research could focus on exploring online augmentation and its integration into model training.
\item Investigating combinations of different augmentation techniques could create more diverse and robust datasets, potentially improving model performance.
\item Examining the effectiveness of augmentation strategies across a wider range of NLP tasks and various model architectures could yield further valuable insights.
\end{itemize}

In conclusion, our study sheds light on the multifaceted nature of text augmentation in NLP. By highlighting the strengths and limitations of various methods, and suggesting future research directions, this work serves as a valuable guide for practitioners and researchers in the field of NLP.

\printcredits

\bibliographystyle{cas-model2-names}

\bibliography{cas-refs}




\end{document}